# Autonomous Underwater Vehicle: Electronics and Software Implementation of the Proton AUV

Vivek Mange[1], Priyam Shah[2], Vishal Kothari[3]

[1]Student, Dept. of Electronics and Telecommunication Engineering, K J Somaiya College Of Engineering, Mumbai, India.

[2,3]Student, Dept. of Computer Engineering, K J Somaiya College Of Engineering, Mumbai, India.

---***---

**Abstract -** *The paper deals with the software and the electronics unit for an autonomous underwater vehicle. The implementation in the electronics unit is the connection and communication between SBC, pixhawk controller and other sensory hardware and actuators. The major implementation of the software unit is the algorithm for object detection based on Convolutional Neural Network (CNN) and its models. The Hyperparameters were tuned according to Odroid Xu4 for various models. The maneuvering algorithm uses the MAVLink protocol of the ArduSub project for movement and its simulation.*

## 1.INTRODUCTION

Recent technological advancement and interest of people towards the Single Board Computers have uplifted and proved to be of great usefulness in the robotics and drone industry. With the cutting edge technology for SBCs and the simulation of the AUVs has led to various emerging communities towards its development such as ArduPilot, being one of the popular ones which comprise of subdirectories like ArduSub, ArduCopter, etc.

Our System consists of Odroid Xu4 as a SBC, which acts as a brain for controlling and communication. The Pixhawk is connected to Odroid via Serial communication and act as a flight controller for maneuvering and simulation. The Tensorflow Object detection API, along with SSD Mobilenet v2 model helped to achieve object detection underwater. The hyperparameter such as learning rate, batch size, number of epochs, etc was tuned according to practical implementation on Odroid. The motion of the system underwater was executed with the help of python package Pymavlink. The simulation was carried out in the SITL and QGC software before real-time testing. This software provided the data of sensors like gyrometer, accelerometer, etc.

Underwater vehicles and technologies like AI have caught people's eye towards this field. Research and development in this field have led to various emerging applications such as cleaning water robots, oil excavation, etc.

## 2. SINGLE BOARD COMPUTER (SBC)

Single Board Computers (SBCs) are high-speed micro-controllers. The SBC is responsible for manipulating different aspects of the system and process the information. Odroid Xu4 and Raspberry Pi 3B are two of the processors on which we worked. Selection of a better processor is based on the processing speed, GPU, external ports, etc for successfully completing the vital tasks within a given time period.

### 2.1  Raspberry Pi 3B

Raspberry Pi 3B is one of the latest models in the raspberry series. RPi 3B is a 1.2Ghz quad-core processor with Broadcom BCM2837 SOC 64bit CPU for fast processing. RPi 3B has a 1GB LPDDR2 RAM as a storage unit for faster data rate with low power consumption. It has an ethernet connection with 100 Mbps transmission but does not allow to power over Ethernet (PoE). The power consumption of RPi 3B is around 260mA in idle condition and up to 730mA, when processed at 400% CPU load. RPi has inbuilt wireless LAN support specified under IEEE 802.11b/g/n band and a Bluetooth module. It has few external ports such as 1x Full-size HDMI port, 4x USB ports with USB 3.0 and an external port for connecting an RPi camera. It has a Micro SD card slot for loading the operating system and storing the data.[4]

### 2.2  Odroid Xu4

Odroid Xu4 is the latest generation SBC. It possesses a faster processing speed with dual quad-core processor. The two processors are Samsung Exynos5422 Cortex[TM]-A15 2Ghz and Cortex[TM]-A7 CPUs with additional GPU support of Mali-T628 MP6. It comprises of 2GB LPDDR3 RAM for storage and high data rate. It has a Gigabit Ethernet to possess a high-speed data transmission. Odroid Xu4 requires high power input up to 5V/4A while initial running, which leads to a high amount of heat dissipation, and to overcome this issue we have an

'External FAN' or a 'Heat Sink' install on the board. It lacks in the in-built assembly of any WiFi or Bluetooth system. It has 1x HDMI display port, 2x USB 3.0 and 1x USB 2.0. It also possesses an Emmc 5.0 module for flash storage.[5]

## 2.3 Analysis of Electronics

Initially, We used RPi 3B as our main processor to perform various tasks such as Maneuvering, Image processing, and Acoustic communication. The RPi was integrated with Pixhawk to command the thrusters for any particular motion resulting in the maneuvering of the system.

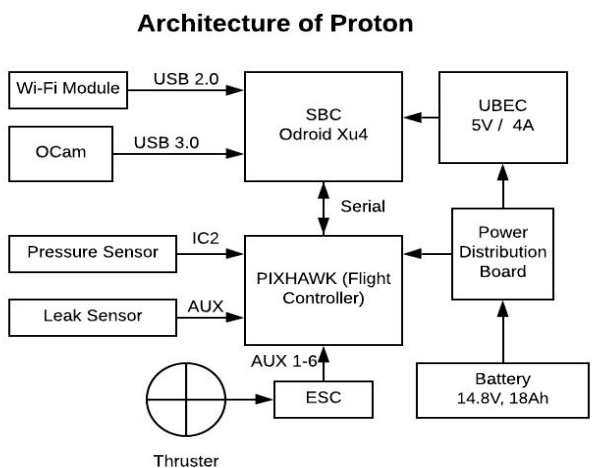

**Fig-1 : Architecture of Proton**

Image Processing that comprises of the object detection system faces few issues while working on RPi as the delay in the output were, fewer frames per second ratio(FPS) and increase in the system temperature. To overcome this issue we upgraded our system to Odroid Xu4 which helped in debugging the issues of the FPS up to 30 times faster than RPi. Odroid surpasses RPi in processing speed as well by providing approximately 7 times faster processing power. Odroid Xu4 is integrated with Pixhawk using MAVLink protocol and the complete architecture is shown in (Fig-1). An external WiFi module and an O-CAM is connected to perform the necessary tasks.

## 3. OBJECT DETECTION

Object detection is technology related to computer vision and image processing that deals with detecting instances of semantic objects of a certain class in a digital image. Object detection is more powerful than classification, it can detect multiple objects in the same image. It also tags the objects and shows their location within the image.[2]

There are various techniques to detect an object in an image, one of them is using neural networks. We have used CNN to detect objects in the image.

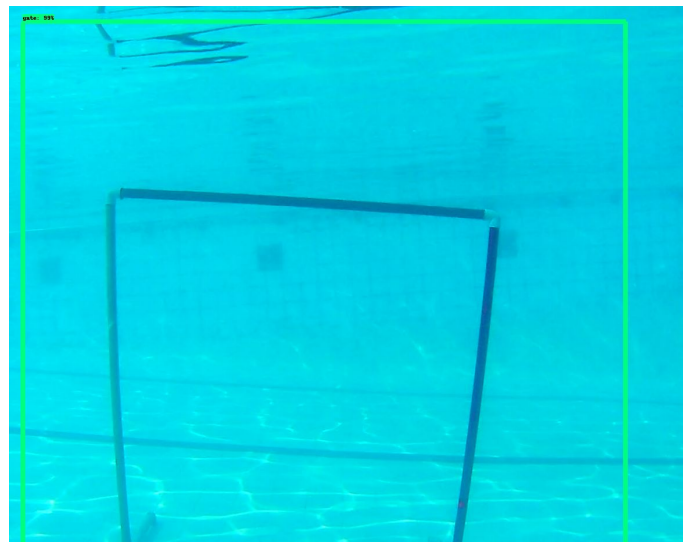

**Fig-2 : Underwater Object Detection**

### 3.1 Convolutional Neural Network (CNN)

The convnet is a special kind of neural networks which contains at least one convolutional layer. A general convnet structure will take an image, then pass it through a series of convolutional layers, followed by pooling and a fully connected layer to output the classification labels.

A neural network consists of several different layers such as the input layer, at least one hidden layer, and an output layer. Neural networks are best used in object detection for recognizing patterns such as edges, shapes, colors, and textures. The hidden layers are convolutional layers in this type of neural network which acts like a filter that first receives input, transforms it using a specific pattern/feature, and sends it to the next layer. With more convolutional layers, each time a new input is sent to the next convolutional layer, they are changed in different ways (fig-3). For example, identifying a blue car, from a given dataset consisting images of cars and bikes, in the first convolutional layer, the filter may identify shape in a region to find a car, and the next one may be able to conclude the color of the car, and the last convolutional layer may classify the car as blue car (Fig-3). Basically, as more and more layers the input goes through, the more sophisticated patterns the future ones can detect.

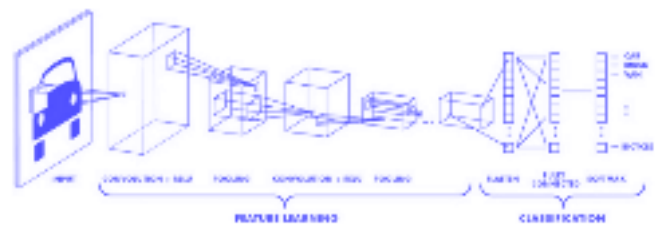

**Fig-3 : CNN**

There are two types of deep neural networks. Base network and detection network. MobileNet, VGG-Net,

LeNet are few examples of base networks. Base network provides high-level features for classification or detection. If you use a fully connected layer at the end of these networks, you get an image classifier. But you can remove the fully connected layer and replace it with detection networks, like SSD, Faster R-CNN, and so on.

### 3.1.1 Faster R-CNN v2

R-CNN stands for Region Convolutional Neural Network. In the object detection, an R-CNN is used to find regions since determining the location of multiple objects is essential to this type of model. An image is split into a lot of different boxes to check if any of them have signs of the object. The algorithm uses region proposal networks (RPN) which ranks the specific regions that are most likely to have the object and this is done using classifier and a regressor (fig-4). A Classifier determines the probability of a proposal having the target object. Regression regresses the coordinates of the proposals.[1]

Performance and precision for an object detection task using Faster R-CNN v2 model, scores over 28 mAP (mean Average Precision) at 58 ms speed on COCO dataset.

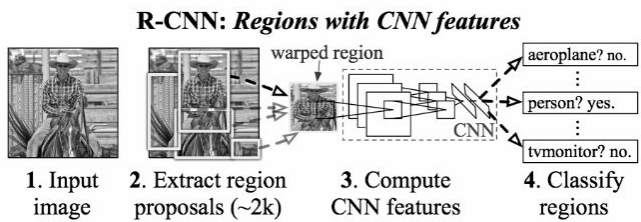

Fig-4 : Faster R-CNN

### 3.1.2 Single Shot MultiBox Detector (SSD) MobileNet v2

**Single Shot:** this means that the tasks of object localization and classification are done in a single forward pass of the network.

**MultiBox:** this is the name of a technique for bounding box regression developed by Szegedy et al.

**Detector:** The network is an object detector that also classifies those detected objects, (fig-5).

Performance and precision for an object detection task using SSD MobileNet v2 model, scores over 22 mAP (mean Average Precision) at 31 ms speed on COCO dataset.

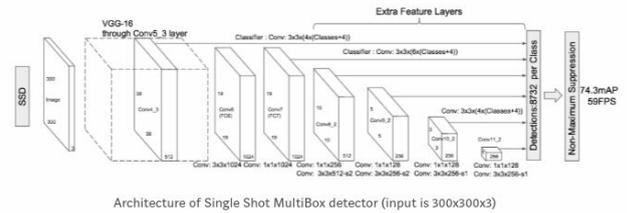

Fig-5 : SSD Mobilenet

The Processing phase is faster in SSD MobileNet v2 model over the Faster R-CNN v2 model in Odroid Xu4.

## 4. HYPERPARAMETER

Hyperparameters are the variables which determine the network structure and how the network is trained. Hyperparameters are set before training and before optimizing the weights and bias for the layers. The training parameters we used were found by practical implementation and tuning of the model. [1]

### 4.1 Learning Rate

Learning rate controls the updation of the weight in the optimization algorithm. Learning rate can be fixed learning rate, gradually decreasing learning rate, momentum-based methods or adaptive learning rates, depending on our choice of optimizers such as Adam, Adagrad, SGD, Ada Della or RMSProp.

Our Training parameters : a) Faster R-CNN v2 model, we used manual_step_learning_rate with initial_learning_rate as 0.0002. b) For SSD Mobilenet v2 model we used exponential_decay_learning_rate with initial_learning_rate as 0.004.

### 4.2 Number of epochs

The number of epochs is the number of times the entire training set pass through the neural network. The variation in epochs can be seen by increasing the number of epochs until we see a small gap between the test error and the training error.

Our Training parameters : The number of epochs = 1 for both models.

### 4.3 Batch Size

Convnet usually prefers small batch size in the learning process. A batch size of 1 to 3 is a good choice to test with Odroid. The convnet is sensitive to batch size.

Our Training parameters : a) Faster R-CNN v2 model, we used batch_size: 1 with momentum optimizer. b) SSD Mobilenet v2 model, we used batch_size: 1 with rms_prop_optimizer.

### 4.4 Activation Function

Activation function introduces non-linearity to the model, which is required for complex functional mappings between the inputs and response variable. Usually, rectifier works well with convnet. Other alternatives are sigmoid, tanh and other activation functions depending on the task.

Our Training parameters : a) Faster R-CNN v2 model, we used softmax/logistic regression. b) SSD Mobilenet v2 model, we used RELU_6.

### 4.5 Weight Initialization

To prevent dead neurons we have to initialize some weight which can be a small random number, but not too small as to avoid zero gradients.

Our Training parameters : a) Faster R-CNN v2 model, we used weight:0.0. b) SSD Mobilenet v2 model, we used weight:0.00004

### 4.6 Dropout for regularization

To avoid overfitting in deep neural networks, we can use some regularization technique and most preferable is the dropout technique. The method simply drops out units in a neural network according to the desired probability or threshold.

Our Training parameters : a) Faster R-CNN v2, we had dropout_keep_probability: 0.1. b) SSD Mobilenet v2, we had dropout_keep_probability: 0.8.

## 5. MANEUVERING

Maneuvering is the movement of the system. There are 5 degrees of freedom available in the System, that provides movements in the respective direction as roll, yaw, heave, dive, sway and surge (Fig-6). We maneuver our system using a python script with the MAVLink protocol functions. The script is uploaded on the Odroid Xu4 and with the integration of Pixhawk, the commands to the particular thrusters are provided with the PWM values. The Pixhawk is connected to ESCs which intakes the input in the form of PWM pulses and the output of ESCs are connected to thrusters to provide the required thrust.

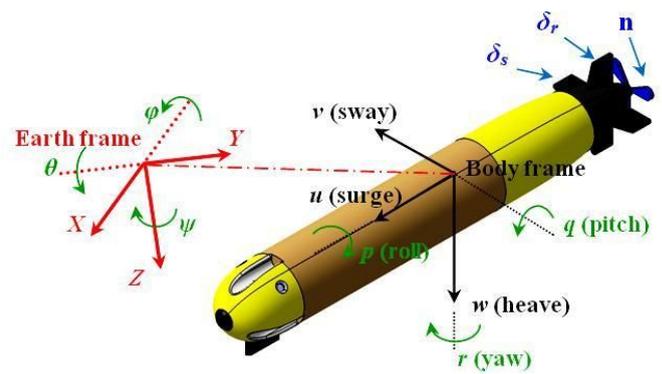

**Fig-6 : Maneuvering Movement**

### 5.1 MAVLink

MAVLink (MICRO AIR VEHICLE COMMUNICATION PROTOCOL) is a binary telemetry protocol. It is specifically designed to establish communication with drones and between its onboard components. MAVLink protocol was designed to create a very lightweight messaging protocol, which used limited system resources and bandwidth. Telemetry data streams delivers a specific message between the Odroid and the pixhawk controller. MAVLink is used in number of autopilots, ground stations and integration APIs.[7]

PyMavlink is a python library for handling ArduSub vehicle communication and maintaining log files using the MAVLink protocol. A python script is used to read the data from different sensors and send commands to an ArduSub vehicle.[7]

### 5.1.1 ArduSub

ArduSub is a sub-project under ArduPilot for underwater projects. It uses the MAVLink protocol ,multiple tools and packages for allowing one to control the system via python scripts based on certain logical interpretation of working environment. ArduSub allows custom package generation as per the developer [6]. Also, adjustment of parameter values can be done based on different geographical variation and application use.

### 5.1.2 Simulator (SITL & QGC)

ArduSub project comes with a Software in the Loop (SITL) simulator to simulate the ArduSub vehicle. SITL allows us to test the code without any hardware implementation and to analyse the desired output. ArduSub along with QGroundControl also offers us GUI support for users on different platforms (Fig-7) [6].

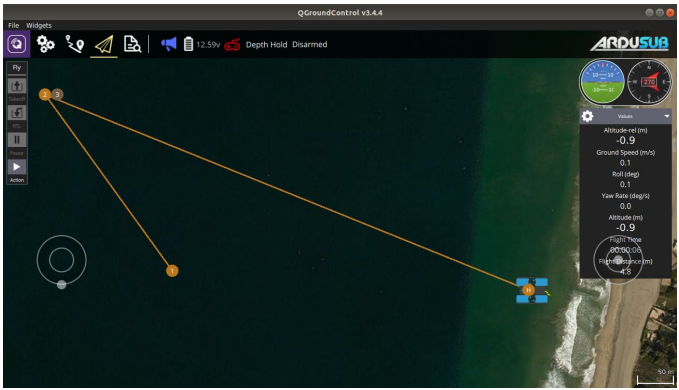

**Fig-7 : Simulation of System using QGC**

## 6. PRACTICAL IMPLEMENTATION

The overall implementation was carried out using Odroid, Pixhawk, electronic equipment and Vehicle.

### 6.1 System Connection and Communication

The Odroid Xu4 with Ubuntu Mate 16.04 was connected through UART0 port to the Telemetry port of Pixhawk (Fig-8). Wifi Module 0 was used to communicate between Odroid and the surface computer. This communication is possible in two ways i) Wired connection ii) Wireless connection. The Wired connection is established through the ethernet connection using Fathom X on both the end as it is used to provide the GUI support on QGC. The wireless connection is established using Putty software. The Pixhawk was configured using QGroundControl software along with Ardusub. The communication between both the devices was established using the python package Pymavlink. The oCam: 5MP is connected to the USB 3.0 of Odroid, which is used for image and video capturing. It provides the frame rate of 1280×720@30fps with 65 degrees Field of vision (FOV).

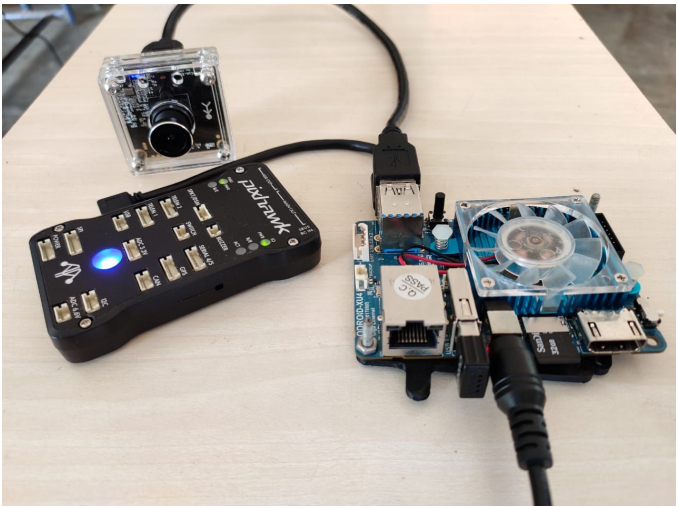

**Fig-8 : Connections**

### 6.2 Training Phase Implementation

The Tensorflow object detection code along with training models and labeled underwater images were trained on a laptop having a graphics card of NVIDIA GTX 1050 Ti [8]. The Faster R-CNN v2 and SSD Mobilenet v2 models were trained on COCO dataset with thousands of classes. Then the models were retrained with 2 new classes namely gate and flare using a large dataset of underwater images respectively. Retraining of previously achieved checkpoints was continued to achieve better results. Using trial and error the inference graphs were generated for each model, whereby the hyperparameters were tuned for various batch size and tested on Odroid.

### 6.3 Testing Phase Implementation

The inference graph along with required code to test was copied on the SD card inserted in Odroid. We took input as Video capture using OpenCV library and oCam camera. Then passed it to our main script for object detection. If the tensorflow detects gate/flare above a particular threshold of 75% then it creates a box around the image with the label of detected object and detection score.[Fig-2] The 4 corner coordinates of this box are then used to find the center of the detected image. The center of the main Video frame is also taken and is subtracted from the detected image center which provides us whether the detected object is on the Left-hand side, Right-hand side, Above, Below or Exact front. Once we have the direction, we now communicate with Pixhawk to forward the required PWM pulses to the ESC. We have separate function codes for each task like Arm/DisArm [On/Off], surge-forward/backward [y-direction movement], sway-right/left [x-direction movement], heave/dive [z-direction], and yaw/roll [rotation]. All the above function codes are called accordingly by the main code to do the necessary movements.

## 7. RESULTS AND ANALYSIS

For better results, we have trained the system with underwater images of the gate and flare with a batch size of 2 and a checkpoint of 25k for SSD mobilenet v2 model. The Threshold was set to a value greater than 75%. The script was tested on underwater videos in real-time.

The system was able to detect objects and make further decision of moving in various directions. The time-lag of processing was around 0.5s on Odroid. We had a dataset of around 5,800 underwater images. Distributed in 80:20 ratio for train and test respectively. About 4640 images for training and 1160 images for testing. The dataset consisted of images labeled gate, flare or both. The training images were taken using GoPro Hero 3 at 60FPS. Various factors were taken into consideration like distance from the object, circular movement, critical sides, partial images, etc. We have also tested the trained Faster R-CNN model and the accuracy was greater than 80%, but the processing time on the embedded devices at real-time was very high around 2s. To carter with high processing speed which is the main aim in real-time, we shifted to SSD Mobilenet models which had comparatively less accuracy, but faster processing speed which resulted in better results.

To stabilize the system various parameters in Pixhawk were manipulated with trial and error. For accomplishing this benchmarks we manipulated various PID parameters such as Roll axis rate with value 0.100, Yaw axis rate with value 0.00 and the Pitch coupling factor with value 1.1 was found to be precise. The different motor direction was changed to generate the perfect torque to stabilize the system.

## 8. CONCLUSION

We conclude that Odroid Xu4 gave us an optimal solution based on its computing power, memory usage, transmission rate, etc over RPi 3B. The SSD Mobilenet v2 model overcame difficulties found using Faster R-CNN v2 model. The tuned hyperparameter values led to better results and accuracy in detection. Maneuvering of the system was successfully done using Pymavlink. For more flexible operations in maneuvering, one can go for the integration of Odroid Xu4 and Arduino Mega.